\documentclass[a4paper, 10pt, conference]{ieeeconf}      

\IEEEoverridecommandlockouts                              

\usepackage[T1]{fontenc}
\usepackage[ansinew]{inputenc}
 \usepackage{multirow}
\usepackage{adjustbox,lipsum}
\usepackage{lmodern} 
\usepackage{amsfonts}
\usepackage{algpseudocode}
\usepackage{savesym}
\savesymbol{AND}
\usepackage{algorithm}
\usepackage{graphicx}
\usepackage{amsmath}
\usepackage{mathtools}

\title{\LARGE \bf
Scheduling Aerial Vehicles in an Urban Air Mobility Scheme}

\author{Emmanouil S. Rigas, Panayiotis Kolios, Georgios Ellinas
\thanks{The authors are with the Department of Electrical and Computer Engineering and the KIOS Research and Innovation Center of Excellence (KIOS CoE), University of Cyprus, emails: \{rigas.emmanouil, kolios.panayiotis, gellinas\}@ucy.ac.cy}
}

\begin{document}

\maketitle
\thispagestyle{empty}
\pagestyle{empty}

\begin{abstract}
Highly populated cities face several challenges, one of them being the intense traffic congestion. In recent years, the concept of Urban Air Mobility has been put forward by large companies and organizations as a way to address this problem, and this approach has been rapidly gaining ground. This disruptive technology involves aerial vehicles (AVs) for hire than can be utilized by customers to travel between locations within large cities. This concept has the potential to drastically decrease traffic congestion and reduce air pollution, since these vehicles typically use electric motors powered by batteries. This work studies the problem of scheduling the assignment of AVs to customers, having as a goal to maximize the serviced customers and minimize the energy consumption of the AVs by forcing them to fly at the lowest possible altitude. Initially, an Integer Linear Program (ILP) formulation is presented, that is solved offline and optimally, followed by a near-optimal algorithm, that solves the problem incrementally, one AV at a time, to address scalability issues, allowing scheduling in problems involving large numbers of locations, AVs, and customer requests.  
\end{abstract}


\section{Introduction}
\label{sec:intr}

Recently, large companies and organizations such as Airbus,\footnote{https://www.airbus.com/innovation/zero-emission/urban-air-mobility.html} Thales,\footnote{https://www.thalesgroup.com/en/canada/press-release/future-air-mobility-relies-thales} and NASA\footnote{https://www.nasa.gov/aeroresearch/one-word-change-expands-nasas-vision-for-future-airspace} initiated plans to transform personal mobility by introducing the concept of Urban Air Mobility (UAM) \cite{STRAUBINGER2020101852}. Urban Air Mobility includes piloted or autonomous electric multicopters, that are used instead of road vehicles to transport people within large cities. UAM advances the current mobility on demand (MoD) concept by adding a third dimension to the options of transportation infrastructure users have. The introduction of UAM is expected to dramatically change the transportation sector, and tackle the ever-increasing traffic congestion problem that affects the quality of life of millions of people that reside in large cities.

Since UAM is still at its infancy, complicated decision-making procedures related to trip scheduling, such as assignment of vehicles to customers, placement of vehicles at specific nodes, the flight levels of the AVs, and the charging of the vehicles, remain unresolved. Optimizing the activities of UAM schemes demands algorithms that can efficiently solve problems that involve several heterogeneous entities (e.g., commuters, multicopter owners, multicopters), that have different goals and needs, operate in dynamic environments (e.g., traffic demands, variable weather conditions, variable consumption and charging characteristics, etc.) and handle uncertainties (e.g., travel requests, energy consumption, etc.). 

A number of related research works exist under the general umbrella of urban air mobility. For example, Patterson et al. \cite{patterson2018proposed} study UAM schemes that carry passengers, focusing on defining a set of requirements for a number of different UAM examples. Additionally, Rothfeld et al. \cite{rothfeld2018agent} propose an extension for the multi-agent transport simulator MATSim that includes UAM and enables a system-wide analysis of an urban air transport scenario, while it considers different aspects such as variations in vertical take-off and landing (VTOL) vehicle properties and the placement of the VTOL infrastructure. Aiming to tackle the crucial problem of multicopter flight level separation, Yang and Wei \cite{yang2020scalable} propose a decentralized message-based computational guidance algorithm, formulating the problem as a multi-agent Markov decision process and solving it with a Monte Carlo tree search algorithm. Focusing on the same problem, Rodionova et al. \cite{9294425} propose a decentralized framework that achieves airborne collision avoidance for multiple vehicles, on-demand. To achieve that, they use signal temporal logic and allow the AVs to execute independently missions that involve spatial, temporal, and reactive objectives. Moreover, the selection of the locations to build the take-off and landing points is crucial in order to maximize the task execution ability of any UAM scheme. In this vein, Willey and Salmon \cite{WILLEY2021102997} study this problem as a modified single-allocation-hub median location problem that incorporates elements of subgraph isomorphism to create structured networks to allow for public transit operations. Given the NP-hardness of the problem, the authors propose several heuristic algorithms that calculate near-optimal solutions having relatively low execution times. 

Additionally, Pradeep and Wei \cite{8569225} focus on minimizing the landing completion time of a given set of electric VTOLs. This is optimized using a heuristic approach combined with two scheduling methods, the first one based on mixed-integer linear programming and the second one on a time-advance algorithm. Further, Haddad et al. \cite{al2020factors} focus on identifying the factors that could affect the adoption of UAM, based on a set of tools and acceptance models \cite{davis1989user}, concluding that safety and trust are of great importance for the adoption of UAM by the wider public. Finally, Afonso et al. \cite{AFONSO2021102688} argue that in order for UAM to become a reality, significant progress on the fully-electric propulsion systems needs to be made, and they propose a methodology to evaluate the environmental footprint of VTOL aircrafts for UAM.

This work investigates the assignment of multicopters to users in a UAM scheme, where the vehicles are located at stations and customers request trips across pairs of stations at particular points in time. The multicopters have a limited range which demands regular battery charging at the stations, while fly level separation is considered to avoid mid-air collisions. In this vein, an optimal offline solution is proposed. Further, given that this solution does not scale to problems with large numbers of AVs, trips, and stations, a near-optimal scalable algorithm is also proposed that incrementally calculates a plan, one vehicle at a time. 

These algorithms are inspired by \cite{RIGAS2018248} and \cite{rigas2020congestion}, where the authors schedule electric vehicles to service customers in mobility-on-demand schemes. Compared to these previous works, the novelty of this work lies in the following: (i) energy consumption is also considered, in relation to the flight level of the AVs, with the objective optimizing not only the number of serviced customers (number of executed tasks), but also the energy consumption of the entire fleet of AVs used in the UAM scheme; (ii) the flying level of the AVs is considered for collision avoidance as well as energy consumption minimization purposes, through the use of an edge intersection mechanism; (iii) the vehicles' charging/discharging model is adapted to the UAM concept by considering not only the horizontal movement of the vehicles, but also the vertical one; and (iv) extensive simulation experiments are performed with regards to the number of serviced customers, energy consumption, AVs' flight levels, and execution time, demonstrating the validity and scalability of the proposed approach.

The rest of the paper is organized as follows: Section~\ref{sec:proDef} provides a formal definition of the problem. Section~\ref{sec:intersec} calculates the set of intersecting edges for a graph embedded on a 2D surface. Section~\ref{sec:mip} formulates the {\it Optimal} algorithm and Section~\ref{sec:incrILP} the {\it Incremental} one. In Section~\ref{sec:eval} a detailed experimental evaluation of the algorithms is performed, while Section~\ref{sec:con} presents concluding remarks as well as future extensions of the proposed work.

\section{Problem Definition}
\label{sec:proDef}
For the UAM setting investigated in this work, customers communicate a priori their traffic demand requests (tasks) to fly between pairs of locations (i.e., takeoff/landing stations) at particular points in time. After all traffic demand requests have been collected, a scheduling algorithm is used to calculate an assignment of AVs to tasks. The aim of this algorithm is to calculate the optimal schedule so as to maximize the executed requests (i.e., executing all tasks may be infeasible due to insufficient resources), while also minimizing energy consumption (in terms of the AVs' flight levels). In this work, it is assumed that all stations are operated by a single UAM company servicing all customers requests, and no competition between different providers exists. It is further assumed that the AVs are using an electric motor which provides them with a limited flying range, thus they have to charge their batteries at the takeoff/landing stations. Note also that one-way transport is assumed and that customers do not stop at in-between locations during a trip (i.e., all customers are serviced via single-hop trips from source to destination). 

Specifically, the flying space is defined as a fully connected graph $G(N,E)$ with nodes $n \in N$ being the takeoff/landing stations of the UAM scheme that accommodates a set of AVs (drones) $a \in A$. Each takeoff/landing station can simultaneously accommodate up to $c_n \in \mathbb{N}$ AVs and it is assumed that each station is equipped with $c_n$ AV  chargers. Time is divided into a set of discrete points $t \in T$, where $T \subset \mathbb{N}$, and each AV has a current location $n^{init}_{a,t}$ which changes when the AV executes a task (i.e., vehicle relocation is not considered). At $t=0$ all AVs are at their initial location and the tasks' execution initiates at $t>1$. At each point in time $t$, each AV has a current location $n_{a,t}$ except from the points in time that this AV is flying from one location to another. Moreover, at each point in time $t$, each AV $a$ has an altitude $\psi_{a,t} \in \{0, \upsilon\}$ where $\upsilon \in \mathbb{N}$, a state of charge $SoC_{t,a} \in \mathbb{N}$, and a remaining range $\tau_t(a)$. Energy consumption follows $con_a + \psi_{a,t} \times hcon_a$; a fixed term $con_a \in \mathbb{R}$ which increases based on $hcon_a \in \mathbb{R}$ during the AV's flight, i.e., depending on the flight level. The charging rate is fixed to $ch_a \in \mathbb{N}$. 

Finally, the set of tasks is defined as $\theta \in \Theta$ and each task is denoted by tuple $\langle n^{start}_\theta, n^{end}_\theta, t_\theta^{start}, \tau_\theta, b_\theta \rangle$. The takeoff and landing stations of task $\theta$ are denoted as $n_\theta^{start}$ and $n_\theta^{end}$, respectively, $t_\theta^{start}$ is the starting point in time for task $\theta$, and $t_\theta^{end} = t_\theta^{start} + \tau_\theta$ is the point in time task $\theta$ should end its execution (i.e., $\tau_\theta$ is the duration of the task). It is assumed, that for a task $\theta$ to be executed, at least one AV $a$ must be located at $n^{start}_a$ at point in time $t_a^{start}$ (i.e., if a task is not executed at its predetermined time it is not rescheduled). The reader should also note that, in this work, the terms drone, multicopter, AV, and vehicle are used interchangeably.  

\section{Finding Intersecting Edges}
\label{sec:intersec}
When designing the graph corresponding to the UAM region of operation, the nodes (station locations) are assumed to be $(x,y)$ points in a 2D Cartesian space and edges interconnect all graph nodes (i.e., a fully connected graph is assumed, with each edge from one node to another being drawn in a straight line following the shortest distance between the two nodes). In order to ensure drone collision avoidance, the edges that intersect (and constraint the drones from flying over these edges simultaneously at the same altitude) must be calculated. Algorithm~\ref{alg:intersect} shown below is used to calculate the set of edges $E^{'}_{e}$ that intersect with edge $e$, $\forall e \in E$. In detail, for every edge $e$, an empty set $E^{'}_{e}$ is initialized, and, for every other edge $d$, it is examined whether it intersects with edge $e$. To do so, a check is performed to ascertain whether the projections of these edges on the $x$- and $y$-axis have a common area in both axes; in that case, edge $d$ is added to set $E^{'}_{e}$, that will eventually include all edges that intersect with edge $e$. 

\begin{algorithm}[!htb]
{\small\begin{algorithmic}[1]
\Require $E$
\ForAll{($e \in E$)}
\State{$E^{'}_{e} \leftarrow \emptyset$} 
	\ForAll{($d \in E: d\neq e$)}
	\\ \{Edge $e$ is defined by $(\alpha,\beta)$ and $(\gamma,\delta)$ and edge $d$ by points $(\epsilon,\zeta)$ and $(\eta,\theta)$\}
	\\	\{Check if edges $e$ and $d$ overlap on the x- AND y-axis\}
	\If {$(\delta-\beta)(\eta-\epsilon)\neq(\gamma-\alpha)(\theta-\delta)$ AND $max(\alpha,\epsilon)\leq \dfrac{\epsilon\dfrac{\theta-\zeta}{\eta-\epsilon}-\zeta -\alpha\dfrac{\delta-\beta}{\gamma-\alpha}+\beta}{\dfrac{\theta-\delta}{\eta-\epsilon}-\dfrac{\delta-\beta}{\gamma-\alpha}}\leq min(\gamma,\eta)$}
		\State{$E^{'}_{e} \leftarrow d$}
	\EndIf
	\EndFor
\EndFor
\end{algorithmic}}
\caption{Intersecting Edge Calculation.}\label{alg:intersect}
\end{algorithm}

\section{Optimal Offline Scheduling}
\label{sec:mip}
An offline algorithm is initially formulated as an ILP (referred to as \textit{Optimal}) that calculates the optimal assignment of AVs to customers. \textit{Optimal} maximizes the number of executed tasks, while at the same time it minimizes energy consumption for the AVs by enforcing flight levels at the lowest possible altitude (Eq.~\ref{obj}). Note that in Eq.~\ref{obj}, $\mu$ is a very small number that is selected so that the second part of the objective function never becomes greater than the first part. This ILP is solved using the IBM ILOG CPLEX 12.10 solver.

In this formulation, five ($5$) decision variables are utilized: (i) $\lambda_{\theta} \in \{0,1\}$, which decides if a task $\theta$ is executed, (ii) $\epsilon_{a,\theta,t} \in \{0,1\}$, which decides if AV $a$ is executing task $\theta$ at point in time $t$, (iii) $\omega_{a,t,n} \in \{0,1\}$, that decides if AV $a$ landed at time $t$ at location $l$, (iv) $h_{a,t} \in \{0, \upsilon\}$, which decides the flight level of AV $a$ at time $t$, and (v) $\phi_{a,t} \in \{0,1\}$, which decides if AV $a$ is recharging at time $t$.\\ 
\\
\textbf{Objective function:}

\begin{equation}
\label{obj}
	max \sum_{\theta \in \Theta}(\lambda_{\theta}) - (\sum_{a \in A}\sum_{t \in T}h_{a,t})\times \mu
\end{equation}

\textbf{Subject to:}

\begin{itemize}
	\item \textit{Task execution constraints:}
\end{itemize}

\begin{equation}
\label{2}
	\sum_{a \in A} \sum_{t^{start}_\theta \leq t< t_\theta^{end}} \epsilon_{a,\theta,t} = \tau_{\theta} \times \lambda_{\delta}, \forall \theta 
\end{equation}

\begin{equation}
\label{3}
	\sum_{a \in A} \sum_{t^{start}_\theta >t , t\geq t_\theta^{end}} \epsilon_{a,\theta,t} = 0, \forall \theta
\end{equation}

\begin{equation}
\label{4}
	\epsilon_{a,\theta,t+1} = \epsilon_{a,\theta,t} \forall a, \forall \theta, \forall t: t^{start}_\theta \leq t < t^{end}_\theta-1 
\end{equation}

\begin{equation}
\label{24}
	\phi_{a,t} \leq \sum_{n \in N}\omega_{a,t,n} \times ch_a, \forall a, \forall t 
\end{equation}

\begin{multline}
\label{5}
	0 \leq SoC_{a,t=0} + \sum_{t'=0}^t \phi_{a,t'} \times ch_a - \\ \sum_{t'=0}^t ((1-\sum_{\theta \in \Theta} \epsilon_{a,\theta,t}) \times con_a + h_{a,t} \times hcon_a) \leq 100, \forall a, \forall t
\end{multline}

The \textit{task execution} constraints ensure that the tasks are properly executed. For each task executed, the AV must fly for the duration of the equivalent trip (Constraint~\ref{2}), and the AV is not flying when it does not execute a task (Constraint~\ref{3}). Additionally, at most one AV can execute a task (Constraints~\ref{4}, \ref{9}). Constraint~\ref{24} ensures that in order for an AV $a$ to be charging, it must be located at one of the stations (i.e., it is not flying), and charging has a rate of $ch_a$. Note that the solver is selecting the points in time to charge considering that the task execution ability of the AV is not reduced. Moreover, Constraint~\ref{5} delimits the state of charge of each AV to always be between the $0-100\%$ range. Thus, no AV will execute a task for which the current state of charge is not enough to provide to the vehicle the necessary range for reaching its intended destination. 

\begin{itemize}
	\item \textit{Temporal, spatial, and routing constraints:}
\end{itemize}

\begin{equation}
\label{7}
	\sum_{n \in N} \omega_{a,t,n} = 1 - \sum_{\theta \in \Theta} \epsilon_{a,\theta,t}, \forall a, \forall t
\end{equation}

\begin{equation}
\label{level1}
|\sum_{n \in N}\omega_{a,t,n}-1| \leq	h_{a,t} \leq \sum_{\theta \in \Theta} \epsilon_{a,\theta,t}\times \upsilon, \forall a, \forall t
\end{equation}

\begin{multline}
\label{level2}
|h_{a,t}-h_{a',t}| \geq \epsilon_{a,\theta,t} + \epsilon_{a',\theta',t} - 1,\\ \forall \theta \in \Theta, \forall t: t^{start}_\theta \leq t < t^{end}_\theta-1 \\ \forall{\theta' \in \Theta: \theta' \neq \theta \& E_{\theta'} \in E_{\theta} \& t^{end}_\theta},\\ \forall{a \in A}, \forall{a' \in A: a'\neq a}
\end{multline}

\begin{equation}
\label{9}
	2 \times \sum_{\theta \in \Theta} \epsilon_{a,i,t_r^{start}} =  \sum_{n \in N} \sum_{t \in T-1} \left| \omega_{a,t+1,l} - \omega_{a,t,n} \right|, \forall{a}
\end{equation}

\begin{equation}
\label{10}
	 \omega_{a,t_{\theta}^{start}-1,n_\theta^{start}} \geq \epsilon_{a,\theta,t_\theta^{start}}, \forall \theta, \forall a
\end{equation}

\begin{equation}
\label{11}
	 \omega_{a,t_{\theta}^{end},l_\theta^{end}}\geq \epsilon_{a,\theta,t_\theta^{end}}, \forall \theta, \forall a
\end{equation}

\begin{equation}
\label{13}
	\omega_{a,t=0,n} = n_a^{start}, \forall a, \forall n
\end{equation}

\begin{equation}
\label{14}
	\epsilon_{a,\theta,t=0} = 0, \forall a, \forall \theta
\end{equation}

\begin{equation}
\label{12}
	\sum_{a \in A} (\omega_{a,t,n}) \leq c_n^{max}, \forall n, \forall t
\end{equation}

The \textit{temporal, spatial, and routing} constraints ensure that the AVs are properly placed over time. In particular, Constraint~\ref{7} enforces all AVs to either be executing tasks or be located at a station. Moreover, together with Constraint~\ref{4} it ensures that at any point in time, each AV executes no more than one task. Further, Constraint~\ref{level1} ensures that when an AV is not flying its flight level is equal to $0$, and when it executes any task its flight level is greater than $0$. Additionally, for each pair of tasks that involve trips over edges that intersect and overlap in time, the flight level of the AVs must be different to avoid mid-air collisions (Constraint~\ref{level2}). Also, Constraint~\ref{9} ensures that no location change for any AV takes place, without this AV executing a task (i.e., AVs cannot change locations if they are not servicing customers). Note that, although this constraint is quadratic due to the absolute value, it is linearized at run time by CPLEX. 

Further, for an AV $a$ to execute a task $\theta$, it must be present at the starting location of the task one point in time before the task execution initiates (Constraint~\ref{10}) and at the destination of the task, once the trip has been completed (Constraint~\ref{11}). In doing so, the capacity of every station is not violated at any point in time (Constraint~\ref{12}). Finally, at time $t=0$ all AVs are not flying and are present at their initial locations (Constraint~\ref{13}. 

\section{Incremental ILP Scheduling Algorithm}
\label{sec:incrILP}
It is well known that ILP problems are NP-hard in the worst case \cite{Hutter201479}. Floudas and Lin \cite{floudas2005mixed} describe an efficient approach to solve large-scale, computationally expensive or even intractable, ILP problems, that is based on the decomposition of these problems into smaller sub-problems that are solved sequentially (usually in much shorter times). Although this technique leads to sub-optimal solutions, it has the advantage of making possible the application of ILP formulations to real-world problems. 

\begin{algorithm}[!]
{\begin{algorithmic}[1]
\Require $A$, $N$, $T$, $\Theta$, $\tau_\theta$, $n_a^{initial}$, $\tau_a^{max}$, $c_n$ and $hIncr$.
\State $execTasks \leftarrow \emptyset$
\For{$\forall a \in A$}
	\State Call $Optimal(\Theta, a)$ 
	\State \textbf{Return} Tasks to be executed by AV $a$ 
	\\ \{Add to the set of completed tasks the ones executed by AV $a$\}
	\State $execTasks \leftarrow execTasks + newExecTasks$
	\\ \{Remove the executed tasks from the set of remaining tasks\}
	\State $\Theta \leftarrow \Theta - newExecTasks$
	\State Update $hIncr$
\EndFor
\State \textbf{Return} $execTasks$
\end{algorithmic}} 
\caption{Incremental ILP Scheduling Algorithm}\label{alg:incrmip}
\end{algorithm} 

Using the decomposition technique, an offline scheduling algorithm is developed, that incrementally calls and solves the \textit{Optimal} algorithm, one AV at a time. In this work, the problem was decomposed in terms of the set of AVs, as in the evaluation of the \textit{Optimal} algorithm this was the dimension of the problem that affected the execution time the most (see Section~\ref{sec:eval}). Regarding the flight level separation, in this approach the {\it Optimal} algorithm is adapted as follows: a table $hIncr_{a,t,\theta} \in \{0, \upsilon\}$ is used, where the first dimension of the table corresponds to the AVs, the second to the points in time, and the third to the task that the AV is executing, while the values in the table correspond to the flight level of the AV. After the execution of the {\it Optimal} algorithm for each AV, the equivalent values are added to the array. Additionally, Constraint~\ref{level2} is reformulated as 

\begin{multline}
\label{level2New}
|h_{a,t}-hIncr_{a',t,\theta'}| \geq \epsilon_{a,\theta,t}, \forall \theta \in \Theta, \\ \forall t: t^{start}_\theta \leq t < t^{end}_\theta-1 \forall{\theta' \in \Theta: \theta' \neq \theta \& E_{\theta'} \in E_{\theta} \& t^{end}_\theta}, \\ \forall{a' \in A: a'\neq a}.
\end{multline}
so that, given that in each execution round the schedule of one AV is calculated, the flight level is checked based on the already existing values of $hIncr$ (i.e., the flight levels of the AVs for which the schedule has already been calculated).

In more detail, as illustrated in Alg.~\ref{alg:incrmip}, the set of completed tasks is initialized to the empty set (line $1$). Then, for each AV and for the remaining tasks, the \textit{Optimal} algorithm is called. This step returns the tasks that are selected to be executed from the current vehicle (lines $3$-$4$). The set of completed tasks is subsequently updated with the newly scheduled ones, while these tasks are also removed from the list of the remaining tasks (lines $5$-$8$). Further, the $hIncr$ table is also updated (line $9$). Once the \textit{Optimal} algorithm has been executed for all AVs, the complete set of executed tasks is returned (line $11$). 

\section{Performance Evaluation}
\label{sec:eval}
In this section, the proposed algorithms are evaluated with regards to a number of performance metrics, in order to demonstrate the validity and scalability of the proposed approach. Specifically, two sets of simulation experiments are performed:
\begin{itemize}
	\item EXP1: Evaluates the {\it Optimal} and {\it Incr} algorithms in terms of the average number of executed tasks (i.e., number of customers serviced), energy consumption and AVs' flight levels. 	
	\item EXP2: Calculates the execution time of the {\it Optimal} and {\it Incr} algorithms and addresses the scalability of the algorithms. 
\end{itemize}

For all experiments, the following settings are used: $30$ points in time, station locations which are points in the Cartesian space, trips and starting times selected from uniform distributions, trip durations between $1$ and $4$ points in time, maximum capacity of each location $c_l^{max} = 3$, energy consumption rate $con_a = 5$ and $hcon_a = 0.1$, and charging rate of $ch_a = 10$. Note that, $con_a$ and $ch_a$ are parameters that depend on the type of AV utilized; in these experiments a homogeneous fleet of vehicles is assumed with the same charging and consumption rates. To execute the aforementioned experiments, a Windows PC is utilized equipped with an Intel i7-4790K CPU and 16 GB of RAM running at 2400MHz. 

\subsection{EXP1: Execution of tasks}
\begin{figure}[!htb]
  \centering
    \includegraphics[scale=0.65]{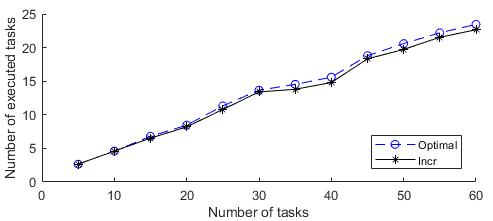}
		\vspace{-5pt}
		\caption{Task execution vs number of tasks ($5$ drones).}
		\label{fig:execTasks}
		\vspace{-5pt}
\end{figure}

\begin{figure}[!htb]
  \centering
    \includegraphics[scale=0.65]{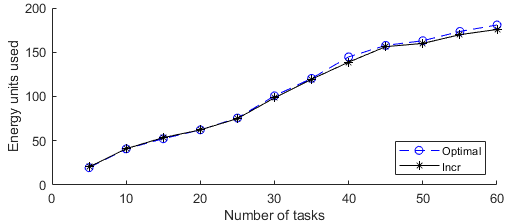}
		\vspace{-5pt}
		\caption{Energy units used vs number of tasks ($5$ drones).}
		\label{fig:execTasksEnergy}
		\vspace{-5pt}
\end{figure}

\begin{figure}[!htb]
  \centering
    \includegraphics[scale=0.3]{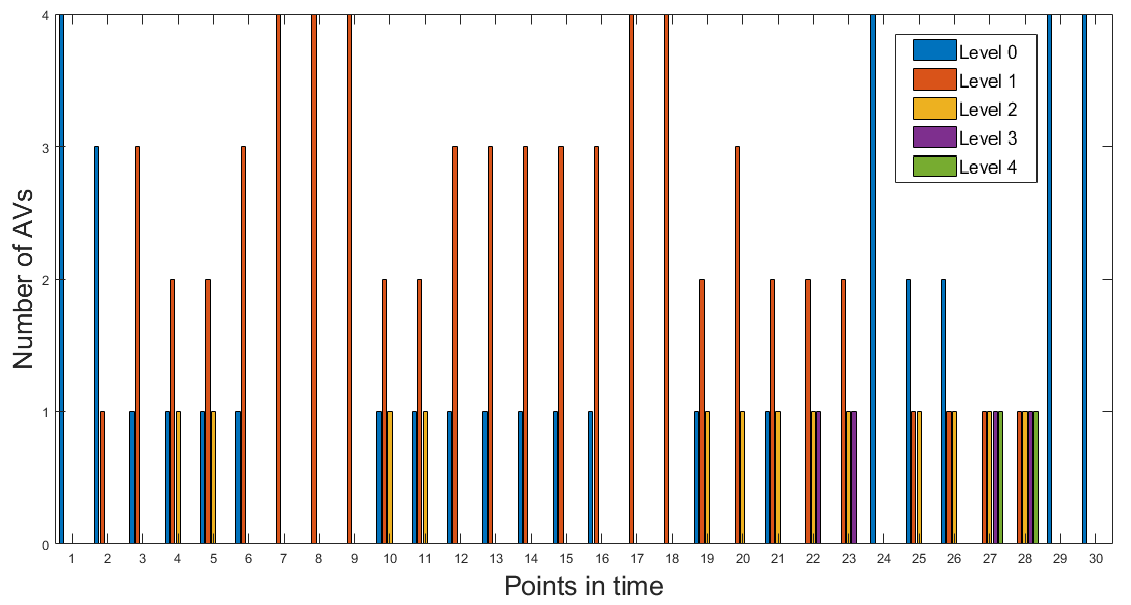}
		\caption{Example execution - flight levels.}
		\label{fig:exLevel}
		\vspace{-5pt}
\end{figure}

\begin{figure}[!htb]
  \centering
    \includegraphics[scale=0.45]{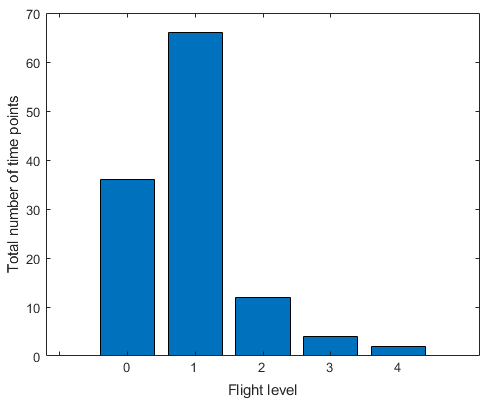}
		\vspace{-5pt}
		\caption{Total number of time points for all AVs at each flight level.}
		\label{fig:exLevelTotal}
		\vspace{-5pt}
\end{figure}

Initially, the average number of customers serviced (i.e., average number of executed tasks) is investigated when using the {\it Incr} algorithm in comparison to the {\it Optimal} one. In a setting with $5$ AVs and up to $60$ tasks, it is observed that {\it Incr} performs close to the optimal, as it is at $96.53\%$ of the {\it Optimal} in the worst case (Fig.~\ref{fig:execTasks}). At the same time, it is observed that when the {\it Incr} algorithm is used, the AVs consume in the worst case $3.31\%$ more energy compared to the {\it Optimal} (Fig.~\ref{fig:execTasksEnergy}), due to the fact that the flight level is not optimized, since each AV's flight level is calculated based only on the solution obtained for the AVs that have already been scheduled when the current AV is considered. In this vein, Fig.~\ref{fig:exLevel} depicts an example execution with $4$ AVs and $50$ tasks and shows the number of AVs flying at each level for each point in time using the {\it Optimal} algorithm. Additionally, Fig.~\ref{fig:exLevelTotal} depicts the total number of points in time the AVs are at one of five possible flight levels (with flight level $0$ denoting that the AV is on the ground). From this figure, it is observed that the majority of the time the AVs fly at the lowest possible altitude (i.e., flight level $1$). This is due to the fact that the higher they fly, the more energy they consume, so the {\it Optimal} algorithm calculates the lowest possible flight levels, given the edge intersecting constraints (i.e., ensuring no mid-air collision). 

\begin{figure}[!htb]
  \centering
    \includegraphics[scale=0.65]{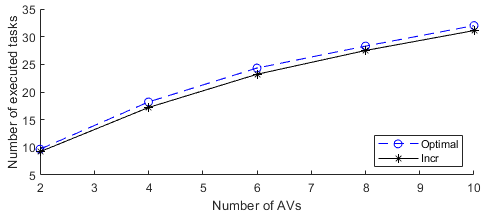}
		\vspace{-5pt}
		\caption{Task execution vs number of drones ($50$ tasks).}
		\label{fig:execDrones}
		\vspace{-5pt}
\end{figure}

\begin{figure}[!htb]
  \centering
    \includegraphics[scale=0.65]{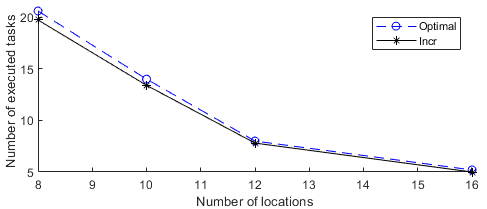}
		\vspace{-5pt}
		\caption{Task execution vs number of locations ($5$ drones, $50$ tasks).}
		\label{fig:execLocs}
		\vspace{-5pt}
\end{figure}

When the total number of tasks is fixed to $50$ but the number of AVs varies (Fig.~\ref{fig:execDrones}), it is shown that the {\it Incr} approach is, in the worst case, at $94.35\%$ of the {\it Optimal}. Finally, when the number of AVs is fixed to $5$ and the number of tasks to $50$, but the number of stations varies, {\it Incr} is at $95.69\%$ of the {\it Optimal} in the worst case (Fig.~\ref{fig:execLocs}). In this experiment, it is observed that the number of tasks that are executed decreases as the number of stations increases. This observation leads to the conclusion that when the number of stations increases, the task execution ability is reduced, as vehicles are less likely to be at locations with demand at the correct point in time. 

\subsection{EXP2: Execution Time and Scalability}
\begin{figure}[!htb]
  \centering
    \includegraphics[scale=0.65]{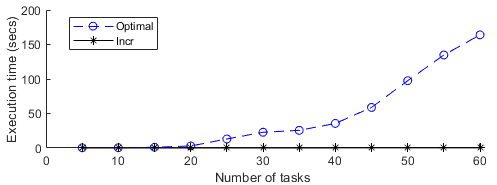}
		\vspace{-5pt}
		\caption{Execution time vs number of tasks ($5$ drones).}
		\label{fig:timeTasks}
		\vspace{-5pt}
\end{figure}

\begin{figure}[!htb]
  \centering
    \includegraphics[scale=0.65]{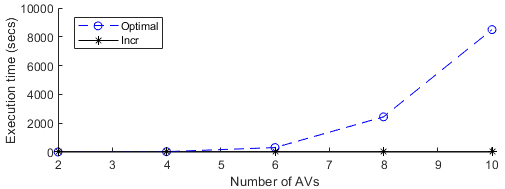}
		\vspace{-5pt}
		\caption{Execution time vs number of drones ($50$ tasks).}
		\label{fig:timeDrones}
		\vspace{-5pt}
\end{figure}

\begin{figure}[!htb]
  \centering
    \includegraphics[scale=0.65]{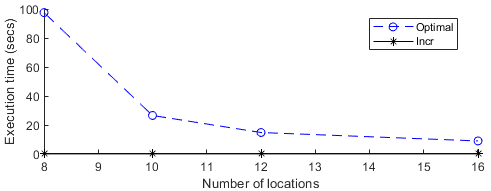}
		\vspace{-5pt}
		\caption{Execution time vs number of stations ($50$ tasks, $5$ drones).}
		\label{fig:timeLocs}
		\vspace{-5pt}
\end{figure}

\begin{figure}[!htb]
  \centering
    \includegraphics[scale=0.65]{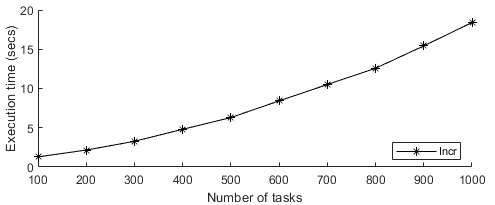}
		\vspace{-5pt}
		\caption{Execution time vs number of tasks ($100$ drones, $8$ stations).}
		\label{fig:incrLarge}
		\vspace{-5pt}
\end{figure}

Execution time and scalability are critical factors concerning the applicability of any scheduling algorithm. In this case, as illustrated in Fig.~\ref{fig:timeTasks}, when the number of AVs is fixed and the number of tasks increases, the execution time of the {\it Optimal} algorithm has a relatively steep increase, as it reaches around $160$ seconds for $60$ tasks, while {\it Incr} has a low execution time, always remaining under $0.5$ seconds even in the larger size problems. Examining the obtained results, it is interesting to note that for more than $50$ tasks the execution time increases with a smaller rate. This can be explained by the fact that after this number of tasks the AVs have almost reached their maximum utilization and the number of additional tasks that can be executed reduces. At the same time, for $50$ tasks, as the number of AVs increases a very steep increase in the execution time of the {\it Optimal} algorithm is observed (for $10$ AVs the execution time is $\sim$ $8000$ seconds), while {\it Incr} retains its low execution time of less than $1$ second (Fig.~\ref{fig:timeDrones}). Moreover, when the number of AVs and tasks is fixed, but the number of locations increases, a gradual decrease in the execution time of both algorithms is observed. This occurs as now fewer tasks are executed, which simplifies the solution to the problem (Fig.~\ref{fig:timeLocs}). Finally, in a larger setting with $100$ AVs and variable number of tasks, it is observed that the execution time of {\it Incr} increases with a relatively low rate (Fig.~\ref{fig:incrLarge}), demonstrating that {\it Incr} can scale to practical-size problems.  

\section{Conclusion}
\label{sec:con}
This work investigates the scheduling of AVs in an Urban Air Mobility scheme, where customer demand is known a priori. The aim is to maximize the number of completed tasks (i.e., number of customers serviced), while also optimizing the flight level of the AVs, in order to avoid mid-air collisions and minimize energy consumption. Two offline scheduling algorithms are proposed, {\it Optimal} and {\it Incr}, with the former calculating the optimal solution, while the latter obtaining near-optimal solutions (in terms of tasks completed and energy consumption), but with increased scalability. Performance results indicate that scalability of {\it Optimal} is constrained by the number of AVs, while {\it Incr}, that solves the ILP formulation sequentially for each AV, can scale to large numbers of locations, AVs, and tasks, while producing results close to the optimal.

Future work includes the development of a heuristic algorithm to solve very large problems, as well as the development of an equivalent online algorithm. Moreover, the relocation of the vehicles in order to decrease idle time and increase the task execution ability will also be considered. 

\section*{Acknowledgment}
This work was supported by the European Union's Horizon 2020 Research and Innovation Programme under Grant 739551 (KIOS CoE) and from the Republic of Cyprus through the Directorate General for European Programmes, Coordination, and Development.


\begin{thebibliography}{1}

\bibitem{STRAUBINGER2020101852}
A. Straubinger, R. Rothfeld, M. Shamiyeh, K-D. Buchter, J. Kaiser, and K.O. Plotner, ``An Overview of Current Research and Developments in Urban Air Mobility - Setting the Scene for UAM Introduction'', \emph{Journal of Air Transport Management}, 87:101852, 2020. 

\bibitem{patterson2018proposed}
M.D. Patterson, K.R. Antcliff, and L.W. Kohlman, ``A Proposed Approach to Studying Urban Air Mobility Missions Including an Initial Exploration of Mission Requirements'', \emph{Proc. AHS International 74th Annual Forum \& Technology Display}, 2018. 

\bibitem{rothfeld2018agent}
R. Rothfeld, M. Balac, K.O Ploetner, and C. Antoniou, ``Agent-based Simulation of Urban Air Mobility'', \emph{Proc. Modeling and Simulation Technologies Conf.}, 2018.

\bibitem{yang2020scalable}
X. Yang, and P. Wei, ``Scalable Multi-Agent Computational Guidance with Separation Assurance for Autonomous Urban Air Mobility'', \emph{Journal of Guidance, Control, and Dynamics}, 43(8):1473--1486, 2020. 

\bibitem{9294425}
A. Rodionova, Y.V. Pant, K. Jang, H. Abbas, and R. Mangharam, ``Learning-to-Fly: Learning-based Collision Avoidance for Scalable Urban Air Mobility'' \emph{Proc. IEEE 23rd International Conference on Intelligent Transportation Systems}, 2020. 

\bibitem{8569225}
P. Pradeep and P. Wei, ``Heuristic Approach for Arrival Sequencing and Scheduling for eVTOL Aircraft in On-Demand Urban Air Mobility'' \emph{Proc. IEEE/AIAA 37th Digital Avionics Systems Conference}, 2018. 

\bibitem{WILLEY2021102997}
L.C. Willey and J.L. Salmon, ``A Method for Urban Air Mobility Network Design using Hub Location and Subgraph Isomorphism'' \emph{Transportation Research Part C: Emerging Technologies}, 125, 102997, 2021

\bibitem{al2020factors}
C. Al Haddad, E. Chaniotakis, A. Straubinger, K. Pl{\"o}tner, and C. Antoniou, ``Factors Affecting the Adoption and Use of Urban Air Mobility'', \emph{Transportation Research Part A: Policy and Practice}, 132:696--712, 2020. 

\bibitem{davis1989user}
F.D. Davis, R.P. Bagozzi, and P.R. Warshaw, ``User Acceptance of Computer Technology: A Comparison of Two Theoretical Models'', \emph{Management Science}, 35(8):982--1003, 1998. 

\bibitem{AFONSO2021102688}
F. Afonso, A. Ferreira, I. Ribeiro, F. Lau, and A. Suleman,``On the Design of Environmentally Sustainable Aircraft for Urban Air Mobility'', \emph{Transportation Research Part D: Transport and Environment}, 91, 102688, 2021 

 

\bibitem{RIGAS2018248}
E.S. Rigas, S.D. Ramchurn, and N. Bassiliades, ``Algorithms for Electric Vehicle Scheduling in Large-scale Mobility-on-demand Schemes'', \emph{Artificial Intelligence}, 262:248--278, 2018. 

\bibitem{rigas2020congestion}
E.S. Rigas and K. Tsompanidis, ``Congestion Management for Mobility-on-Demand Schemes that use Electric Vehicles'', in \emph{Multi-Agent Systems and Agreement Technologies}, Springer International Publishing, 2020. 

\bibitem{floudas2005mixed}
C.A. Floudas and X. Lin, ``Mixed Integer Linear Programming in Process Scheduling: Modeling, Algorithms, and Applications'', \emph{Annals of Operations Research}, 139(1):131--162, 2005. 

\bibitem{Hutter201479}
F. Hutter, L. Xu, H.H. Hoos, and K. Leyton-Brown, ``Algorithm Runtime Prediction: Methods and Evaluation'', \emph{Artificial Intelligence}, 206:79--111, 2014. 

\bibitem{BRAFMAN201352}
R.I. Brafman and C. Domshlak, ``On the Complexity of Planning for Agent Teams and its Implications for Single Agent Planning'', \emph{Artificial Intelligence}, 198:52--71, 2013. 

\end{thebibliography}
\end{document}